\documentclass[twocolumn,letterpaper]{article}
\usepackage[T1]{fontenc}
\usepackage[utf8]{inputenc}
\usepackage{lmodern}
\usepackage{geometry}
\usepackage{amsmath,amssymb,amsfonts,amsthm}
\usepackage{bm}
\theoremstyle{plain}

\theoremstyle{definition}

\theoremstyle{remark}

\usepackage{algorithm}
\usepackage{algpseudocode}
\usepackage{enumitem}
\usepackage{booktabs}
\usepackage{makecell}
\usepackage{adjustbox}
\usepackage{ragged2e}
\usepackage{authblk}
\usepackage{url}
\usepackage{xcolor}
\usepackage[numbers,sort,round]{natbib} 
\usepackage{hyperref}
\hypersetup{colorlinks=true, linkcolor=blue, citecolor=blue}
\usepackage[capitalize]{cleveref}

\algrenewcommand{\algorithmicindent}{1.5em}
\begin{document} 
	
	\title{Enhancing Imbalanced Electrocardiogram Classification: A Novel Approach Integrating Data Augmentation through Wavelet Transform and Interclass Fusion}
	
	\author{
		Haijian Shao\thanks{Corresponding Author: \href{mailto:jsj\_shj@just.edu.cn}{jsj\_shj@just.edu.cn}}, Wei Liu, Xing Deng, Daze Lu\\
		School of Computer, Jiangsu University of Science and Technology, Zhenjiang 212003, China\\
		Department of Electrical and Computer Engineering, University of Nevada, Las Vegas, 89115, USA
	}
	
	\maketitle 
	
	\begin{abstract}

\noindent Imbalanced electrocardiogram (ECG) data hampers the efficacy and resilience of algorithms in the automated processing and interpretation of cardiovascular diagnostic information, which in turn impedes deep learning-based ECG classification. Notably, certain cardiac conditions that are infrequently encountered are disproportionately underrepresented in these datasets. Although algorithmic generation and oversampling of specific ECG signal types can mitigate class skew, there is a lack of consensus regarding the effectiveness of such techniques in ECG classification. Furthermore, the methodologies and scenarios of ECG acquisition introduce noise, further complicating the processing of ECG data. This paper presents a significantly enhanced ECG classifier that simultaneously addresses both class imbalance and noise-related challenges in ECG analysis, as observed in the CPSC 2018 dataset. Specifically, we propose the application of feature fusion based on the wavelet transform, with a focus on wavelet transform-based interclass fusion, to generate the training feature library and the test set feature library. Subsequently, the original training and test data are amalgamated with their respective feature databases, resulting in more balanced training and test datasets. Employing this approach, our ECG model achieves recognition accuracies of up to 99\%, 98\%, 97\%, 98\%, 96\%, 92\%, and 93\% for Normal, AF, I-AVB, LBBB, RBBB, PAC, PVC, STD, and STE, respectively. Furthermore, the average recognition accuracy for these categories ranges between 92\% and 98\%. Notably, our proposed data fusion methodology surpasses any known algorithms in terms of ECG classification accuracy in the CPSC 2018 dataset.

	\end{abstract}
	
	
\section{Introduction}
Medical record data utilized for disease diagnosis and prediction often presents a significant challenge due to the inclusion of infrequent, yet crucial, disease samples. These samples typically exhibit a limited occurrence compared to the more prevalent and frequently observed disease samples. Consequently, it becomes indispensable to address this matter to ensure precise disease diagnosis. Effectively managing this challenge holds paramount importance for both healthcare professionals and researchers, as it can substantially influence the quality and dependability of disease prediction models. Arrhythmias manifest as a prevailing manifestation of cardiovascular disease, underscoring the importance of accurate categorization and identification. With cardiovascular disease surpassing all other causes of human mortality, medical practitioners rely on electrocardiograms (ECGs) to discern arrhythmia disorders. ECGs, which depict the fluctuations in electrical potential induced by cardiac contractions, have experienced a substantial surge in usage due to heightened public health awareness and the growing popularity of home electrocardiogram testers. Nevertheless, constrained medical resources render manual diagnosis of the vast volumes of ECG data impracticable, thus emphasizing the indispensability of automatic arrhythmia classification. The intrinsic electrical impulses of the heart, captured through ECGs, can be employed for arrhythmia diagnosis. Given the preeminent status of cardiovascular disease as a leading cause of human mortality, it has become imperative to develop technologies that effectively detect and prevent it. Arrhythmia, characterized by irregular heartbeats resulting from disruptions in cardiac electrical conduction or origination, serves as a common indicator of cardiovascular disease \cite{ref1, ref2, ref3}. Arrhythmias, whether occurring in isolation or in conjunction with other cardiovascular conditions, can prove fatal in severe cases, thereby highlighting the importance and complexity of promptly identifying and classifying arrhythmia disorders. In routine diagnostic procedures, a comprehensive evaluation is typically based on the patient's medical history, with the physician personally analysing the patient's electrocardiogram (EKG) readings. The introduction of Holter ECG technology has established continuous monitoring of human ECG signals as a standard medical practice, acknowledging the sporadic and intermittent nature of arrhythmia disorders \cite{ref4}. While this method provides physicians with abundant ECG data, relying solely on physician interpretation could result in significant depletion of medical resources and lead to physician fatigue, potentially leading to errors in decision-making.

\begin{figure*}[h]
	\centering
	\includegraphics[width=0.9\linewidth]{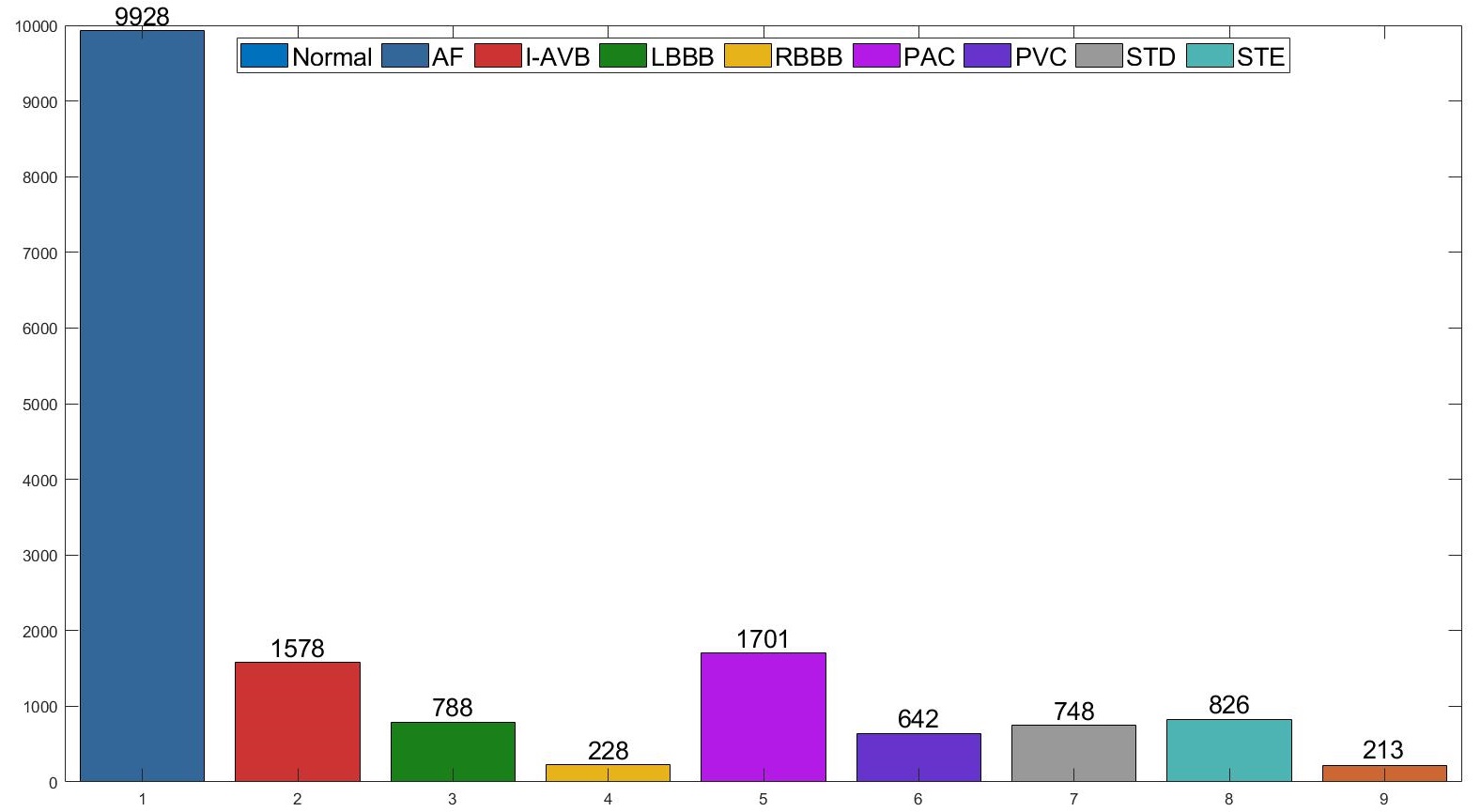}
	\caption{The classes in CPSC 2018 dataset after pre-processing}
	\label{fig1}
\end{figure*}

While significant progress has been made in ECG signal detection, there is still a major challenge that hinders further improvement in accuracy: signal noise and the profound class- imbalance inherent in ECG data sets essential for model training, validation and deployment. In literally every ECG dataset, some classes of ECG signals have significantly fewer samples than others, as exemplified in the China Physiological Signal Challenge 2018 (CPSC 2018) dataset \cite{ref5}. Of its total 16,652 ECG recordings, the majority samples fall into the Normal category, with a staggering 9928 instances, while STE (ST Elevation) and LBBB (Left Bundle Branch Bloc) classes have the smallest numbers of samples, with only 213 and 228 instances, respectively (Figure \ref{fig1}). Such huge gap in imbalance apparently has negative impact on the performance of ECG models, as it may become biased towards the majority class. and the model may struggle to generalize well and accurately classify instances from the underrepresented classes, leading to poor performance in detecting the minority classes.

Owing to the influence exerted by the two primary factors, namely the disparate distribution of categories imbalance and the presence of noise, the prevailing apogee of analytical outcomes, even predicated upon the elucidated performance indices of F1, Faf, Fblock, Fpc, and Fst as officially divulged \cite{ref5}, steadfastly persists at 0.837, 0.933, 0.899, 0.847, and 0.779, correspondingly. These exemplar metrics substantiate the commendable efficacy of the acquired results, thereby endowing the actual analysis with invaluable discernments. Such findings corroborate the fact that an approximate aggregate of 959 samples continues to endure misclassification, consequently implicating a momentous divergence from the attainment of utmost precision and unwavering dependability within the current prognostications. This paper primarily addresses the challenge of the 9-category problem, widely acknowledged for its inherent complexity. The datasets utilized for this study are presented in Table \ref{tab1}. The classes in CPSC 2018 dataset after pre-processing is shown in Figure \ref{fig1}.

\begin{table*}
	\centering
	\caption{Comparing the distribution of data before and after preprocessing}
	\label{tab1}
	\begin{tabular}{lcc}
		\toprule
		Type & Recording & Recording (Pre-processing)\\
		\midrule
		Normal&918&9928\\
		Atrial fibrillation (AF)&1098&1578\\
		First-degree atrioventricular block (I-AVB)&704&788\\
		Left bundle branch block (LBBB)&207&228\\
		Right bundle branch block (RBBB)&1695&1701\\
		Premature atrial contraction (PAC)&556&642\\
		Premature ventricular contraction (PVC)&672&748\\
		ST-segment depression (STD)&825&826\\
		ST-segment elevated (STE)&202&213\\
		Total&6877&16652\\
		\bottomrule
	\end{tabular}
\end{table*}

To address dataset imbalance problems, several techniques have been employed, including,

($i$) oversampling the minority class \cite{ref6}, where synthetic samples are synthetized for the minority class to balance the distribution;

($ii$) undersampling the majority class \cite{ref7}, where samples are randomly removed from the majority class to create a more balanced dataset;

($iii$) class weighting \cite{ref8}, where the minority class during training is assigned higher weights to give it more importance;

($iv$) data augmentation \cite{ref9}, where new samples are created by applying transformations or perturbations to existing data, particularly for the minority class.

Considering the inherent randomness of the undersampling rules \cite{ref10,ref11,ref12} will be resulting from optimal division of data sets can be challenging. Conversely, the class-weighted strategy involves adjusting the weights of categories, such as elevating the weights of underrepresented classes, to mitigate the impact of imbalanced datasets on classification accuracy.

Another salient concern worthy of consideration is that, during the process of ECG signal acquisition, power interference, motion artifacts, myoelectric activity interference, and baseline drift, have the capacity to introduce noise to the ECG recording. Essentially, noise can perturb the morphology, attenuate the amplitude, obnubilate the margins, and even engender spurious ECG waveform characteristics. Filtering and suppression techniques are frequently deployed to mitigate the impact of noise on ECG signal. For instance, low-pass filters can be employed to eliminate high-frequency noise in ECG \cite{ref13,ref14}. To discern and suppress noise components, suppression techniques avail themselves of signal processing algorithms such as adaptive filtering \cite{ref15} and wavelet transform \cite{ref16}.

While remarkable progresses in dataset imbalance and signal denoising have been made separately to help improve the ECG analysis and classification, there is a pressing need for further improvement to enable more precise diagnosis and monitoring of cardiac conditions by addressing these two issues simultaneously. In this paper, we propose a novel method using wavelet transform and deep learning. Specifically, this method picks up one category with a small number of samples, and this sample size is then chosen as the threshold that is applied to screen out the samples from other categories but of equal size. Next, the feature library of the training set is averaged out in the frequency domain to obtain the feature library for the test set. The original training and test sets are then fused with their respective feature libraries, resulting in a new training set and test set that effectively expand the number of samples while minimizing feature loss. Experimental results demonstrate that this method significantly improves the classification performance of benchmark models, such as VGG16, LeNet-5, Inception, and LSTM, on the public dataset CPSC2018, and these experiments also confirm robustness of the proposed method when dealing with noisy ECG for classification.

Researchers and clinics are increasingly employing computer-interpreted classification ECG for diagnosing and monitoring patients’ cardiac health. Classification techniques for arrhythmias can be categorized into traditional machine learning methods and deep learning methods concerning how features are acquired \cite{ref17,ref18}. Traditional machine learning methods create various features pertaining to the diseases of interest and these derived features are subsequently employed by an appropriate classifier to classify the diseases. However, one big disadvantage of machine learning based ECG classification methods is that they rely on costly, error-prone manual feature extraction \cite{ref19}. ECG signal detection and automated processing using deep learning techniques have revolutionized the analysis and interpretation of ECGs \cite{ref20}. The performance of a deep learning ECG model is largely dependent on how well it is trained, and as so, the quality of the ECG data set employed for training purposes has a serious implication on the model validation and deployment. Unfortunately, all the known ECG data sets by nature have imbalanced data, meaning that there is a significant difference in the number of samples between classes. When classes are imbalanced, an ECG model may exhibit biased behavior, favoring the majority class and neglecting the minority class. In this section, we will survey the techniques employed to combat this class-imbalance problem. Besides class-imbalance, the presence of noise in ECG signals, such as baseline wander, muscle artifacts, and powerline interference, can affect the performance of models by introducing unwanted variations and obscuring important cardiac information. In this section, we will also survey various techniques for noise reduction or removal noise in the context of ECG signal detection.

\subsection{Class-imbalance-related approaches for ECG analysis}
The efficacy of undersampling for augmenting sample diversity remains uncertain, and its employed methodology entails notable limitations. Moreover, in scenarios where the disparity in sample selection quantities across classes is minimal, even a marginal increase in the maximum number of samples relative to the smallest class might not yield significant improvements in the classification performance of the model. Furthermore, a substantial influx of repeated inputs can precipitate network weight degradation, as substantiated by empirical analyses. Jiang et al. \cite{ref21} introduced three novel methods, BLSM, CTFM, and 2PT, based on CNNs, to address the problem of class-imbalance in ECG datasets. The experiments conducted using these methods demonstrated improved accuracy in heartbeat detection and classification, indicating their efficacy in addressing the class-imbalance in ECG signals. Similarly, Qiu et al. \cite{ref22} proposed a data-augmentation-based approach to balance the imbalanced ECG datasets. This method involved using normal ECG signals to balance the data among different categories, which resulted in improved performance. The Multi-Feature Transformer (MF-Transformer) was designed for the classification of 12-lead ECG signals, and experiments demonstrated its competitive predictions on five ECG categories. Additionally, Zhang et al. \cite{ref23} developed an active balancing mechanism (ABM) based on entropy information to balance imbalanced data. This mechanism evaluates sample information and retains only valuable samples of the majority class, achieving undersampling. Experiments demonstrated that the proposed ABM, when combined with support vector machines and MCNNs, achieved the best accuracy of 92.31\% and 98.46\%, respectively, even at an imbalance rate of 5. Adib, E. et al. \cite{ref24} utilized generative adversarial networks (GANs) to augment imbalanced ECG datasets, resulting in significant improvements in arrhythmia classification, even in the minor classes. Additionally, deep learning networks have demonstrated stronger nonlinear fitting abilities, which have better effects in identifying single-lead, multi-class, and unbalanced ECG datasets \cite{ref25,ref26}. Hatamian, FN et al. \cite{ref27} investigated the impact of oversampling, Gaussian mixture models (GMMs), and GANs to solve class imbalance problems in ECG datasets. The researchers quantitatively and qualitatively compared their methods with state-of-the-art approaches, demonstrating the performance of different types of methods based on the F1-score. Li et al. \cite{ref28} proposed a novel approach to classifying imbalanced datasets based on knowledge transferred from supervised source signals to unsupervised signals from target users. The experiments conducted on the MIT-BIH Arrhythmias Dataset demonstrated the superiority of the proposed approach over benchmark methods. Finally, Lannoy et al. \cite{ref29} proposed a classifier to identify arrhythmia patterns and pathological heartbeats in ECG signal datasets, addressing both time dependences between observations and strong class imbalance problems.

\subsection{Noise robustness-related approaches for ECG analysis}

ECG signals are millivolt-level bioelectric signals that capture the depolarization of cardiomyocytes and the associated bioelectric activity. In numerous practical applications, ECG signals may become corrupted by various types of noise. For instance, during signal acquisition, baseline drift noise, EMG interference, and power frequency interference can interfere with the ECG signal. Typically, low-pass filtering methods are used for noise suppression of ECG signals, which can effectively filter out additive white noise. However, this method may also attenuate the amplitude of various waves in the ECG signal and cause signal distortion.

In recent studies, Rahman et al. \cite{ref30} investigated the robustness of statistical signal quality indices (SSQIs) against different window sizes across diverse datasets. The authors noted that fluctuations may occur if the adaptive filter threshold is not considered. Additionally, Chatterjee et al. \cite{ref31} evaluated several ECG denoising approaches using widely used databases such as MIT-BIH, PTB, and QT and developed benchmark filters for ECG denoising. Their findings revealed that GAN, the new MP-EKF, DLSR, and AKF are effective in reducing muscular artifacts. GAN, on the other hand, proved to be the best denoising solution for eliminating electrode motion artifacts and base-line drift. DLSR and EWT demonstrated efficacy in removing power-line interference. In detecting QRS waves in ECG signals, the typical three-step process involves bandpass filtering, nonlinear transformation, and rule-based QRS wave detection. However, fixed-band filtering in these methods is not suitable for creating a high-precision classification model for ECG signals. To address this limitation, the wavelet transforms modulus maximum approach can be utilized to cancel out noise in the signal \cite{ref16}. However, this approach involves a significant computational burden, an unstable calculation process, and delayed convergence. Furthermore, ECG signals can exhibit substantial variations both among individuals and over time. For instance, some ECG signals may display sharp peaks in their QRS waves and small amplitudes for the Q and S waves. These variations in waveform morphology can be caused by differences in the anatomical structure of the heart, such as variations in ventricular size or shape. Additionally, the electrical conductivity of the cardiac tissue, which affects the propagation of the electrical impulses, can contribute to the observed variations in the ECG waveform. In the presence of strong noise, the wavelet space adaptation method may result in the Gibbs oscillation phenomenon at the Q and S waves, causing signal distortion and eliminating the geometric properties of ECG signals \cite{ref32,ref33}.

The subsequent sections are organized as follows. Section 2 presents an overview of the approaches specifically tailored to address class imbalance in the analysis of electrocardiogram (ECG) data. The proposed method to deal with ECG class imbalance and noise is detailed in Section 3. In Section 4, the proposed algorithm is applied to classify CPSC2018, and the results are reported, and compared against other known results. Finally, Section 5 concludes the paper.

\section{Deep Learning Classification Approach for Class-imbalanced ECG Signals}
As alluded before, the ECG dataset presents a challenge in this regard due to class imbalance, whereby some categories have a significantly larger sample size than others. The minority classes, with smaller sample sizes, contrast with the majority classes, which feature a greater number of samples. Actually, class imbalance is common in many real-world applications, including medical record data for disease diagnosis and prediction, where rare but critical disease samples are scarce compared to normal or common disease samples. To optimize classification accuracy, the in-class distance within each category in the labeled dataset should be moderate, while the inter-class distance between non-homogeneous categories should be large. This strategy will help maintain a consistent feature distribution and ensure a negligible difference of misclassification across categories.

\subsection{Strategies to reduce class imbalance}
\begin{figure*}[h]
	\centering
	\includegraphics[width=\linewidth]{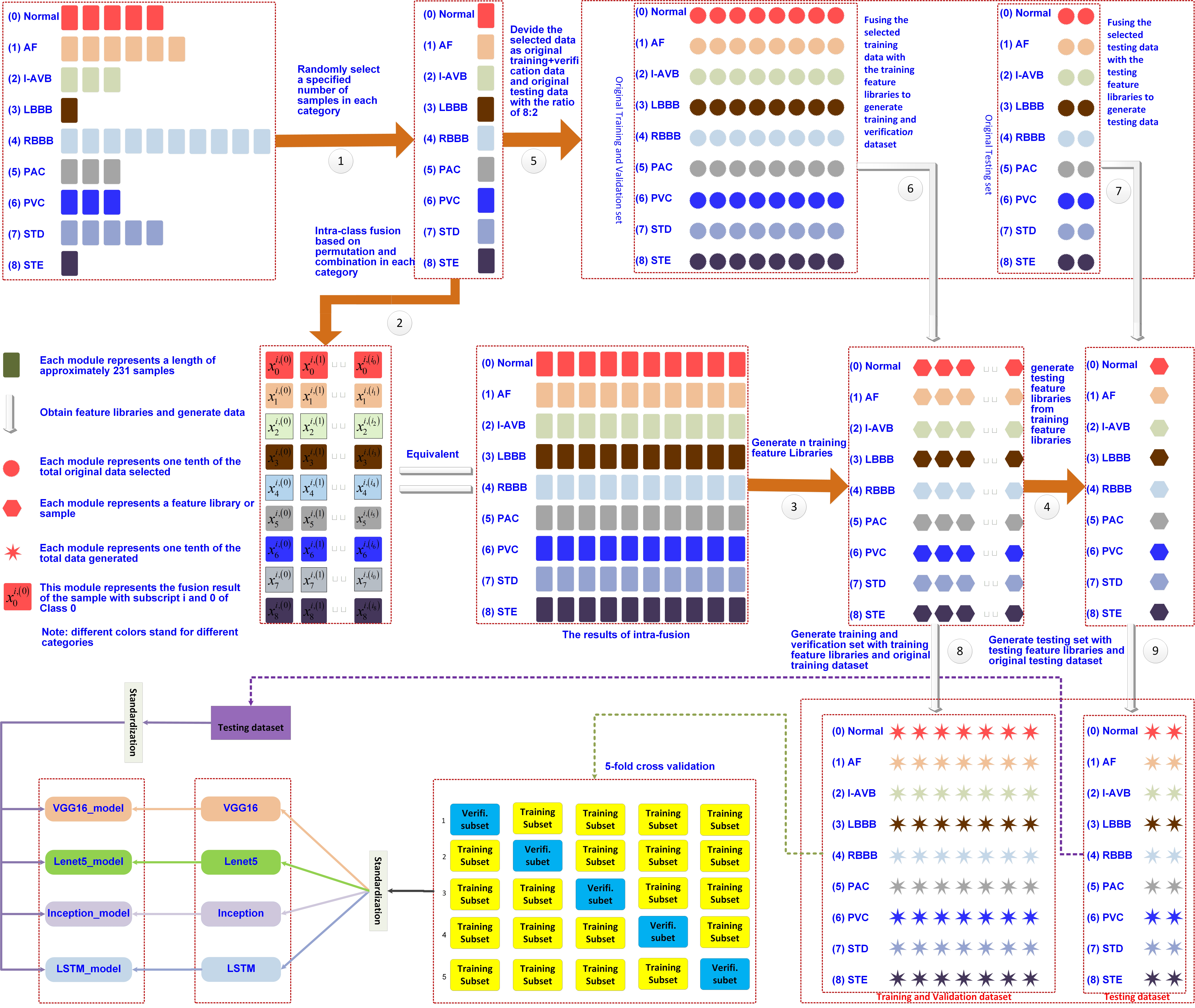}
	\caption{The scheme that handles class imbalance}
	\label{fig2}
\end{figure*}

Class imbalance poses significant challenges in classification tasks. To overcome this issue, a data processing diagram for handling class imbalance has been proposed, as illustrated in Figure \ref{fig2}. Firstly, the wavelet transform is utilized to perform inter-class fusion, which generates the training feature library and the test set feature library. Subsequently, the original training and test data are combined with the training set feature database and the test set feature database, respectively, to generate the new training and test data. Finally, the newly generated training and test data are applied for training and testing purposes.
Figure \ref{fig2} displays modules of different colors, each representing one of nine distinct ECG signals. The quantity of modules is indicative of the data length within their respective categories, signifying that a higher module count corresponds to a larger proportion of data. Notably, RBBB boasts the largest number of modules, while STE has the minimal amount of data.

The procedural steps in Figure \ref{fig2} unfold as follows: in step 1, a specified number of samples are randomly selected within each category. Step 2 involves intra-class fusion through permutation and combination for each category. The results of this intra-fusion process are harnessed to create $n$ training feature libraries in step 3, with testing feature libraries derived from these in step 4. Simultaneously, the data selected in step 1 serves as both the original training+verification and testing data, partitioned in an 8:2 ratio. In step 6, the selected training data, coupled with the training feature libraries, generates the training and verification dataset. Correspondingly, the testing data in step 7, along with testing features, is fashioned. The training and verification set with training feature libraries and the testing dataset with testing feature libraries are showcased in steps 8 and 9, respectively. During the testing stage, four deep learning models—VGG16, Lenet5, Inception, and LSTM—are employed to extract features for classification. In the training and verification phase, the 5-fold cross-validation method is employed to rigorously assess the model's performance and generalization capabilities. The primary objective of 5-fold cross-validation is to furnish a robust evaluation by partitioning the available data into five subsets or folds for comprehensive analysis.

\subsection{Wavelet-based data cleaning and fusion method}

The class with the smallest sample size undergoes initially a cleansing process, after which the number of samples remaining is designated as the threshold value \(n\). For other categories with sample sizes exceeding this threshold, \(n\) samples are randomly selected from the cleansed data to serve as the data for the current category in question. The cleaning process is delineated in Algorithm~\ref{alg1}.

\begin{algorithm}
	\caption{Data cleansing process}
	\label{alg1}
	\begin{algorithmic}[1]
		\Procedure{Input: }{$x$}\Comment{numpy array of a sample}\\
		{$PCA\left( {x,\left( {12,5000 - x.shape\left[ 1 \right]} \right)} \right)$}\Comment{To reduce the dimension of $x$ and obtain the new data ${x_{add}}$ of dimension  $\left( {12,5000 - x.shape[1]} \right)$}\\
		{Output: }{${x_{cleaned}}$}\\
		\quad \textbf{if}{ $x.shape[0] \ne 12$ $\&$ $x.shape[1] \le 2500$}
		\quad \State {$x$ will not be selected}\\
		\quad \textbf{else}\\
		\quad \quad \textbf{if}{ $x.shape[1] \ge 5000$}
		\quad \quad \State $x \gets x\left[ {:,0:5000} \right]$\\
		\quad \quad \textbf{else}
		\quad \quad \State ${x_{add}} \gets PCA\left( {x,\left( {12,5000 - x.shape[1]} \right)} \right)$
		\quad \quad \State $x \leftarrow np.concatenate\left( {\left( {x,{x_{add}}} \right),axis = 1} \right)$\Comment{np.concatenate is used to calculate the numpy array}
		\State \textbf{return} ${x_{cleaned}}$\Comment{The cleaned signal}
		\EndProcedure
	\end{algorithmic}
\end{algorithm}

The presented algorithm \ref{alg1}, designed for data cleansing, takes a numpy array $x$ as input, representing the ECG signals. Employing Principal Component Analysis (PCA), the algorithm reduces the dimensionality of the input array to obtain a new dataset denoted as ${x_{add}}$, with dimensions $\left(12, 5000 - x.shape[1]\right)$. The algorithm selectively processes the input array based on specified conditions: if the number of rows in $x$ is not equal to 12, or the number of columns is less than or equal to 2500, the array is excluded. However, if the number of columns in $x$ exceeds 5000, only the first 5000 columns are retained. If the number of columns is less than 5000, PCA is applied again to obtain ${x_{add}}$, and the cleaned signal ${x_{cleaned}}$ is derived by concatenating the original array $x$ and ${x_{add}}$ along the columns. The resultant ${x_{cleaned}}$ represents the cleaned signal ready for further analysis and processing.

\subsection{Wavelet-based feature fusion method}

The Wavelet-based feature fusion procedure described in this section consists of three stages, followed by the application of the training set's feature base.

Step 1) The selected sample  will perform a two-by-two fusion operation, and n samples in class $k$ are represented as Equ. \ref{eq1}.
\begin{equation}
	\label{eq1}
	{x_{sel}} = \left\{ {x_k^0,x_k^1,x_k^2,...,x_k^{n - 1}} \right\}
\end{equation}

\begin{figure*}[h]
	\centering
	\includegraphics[width=0.6\linewidth]{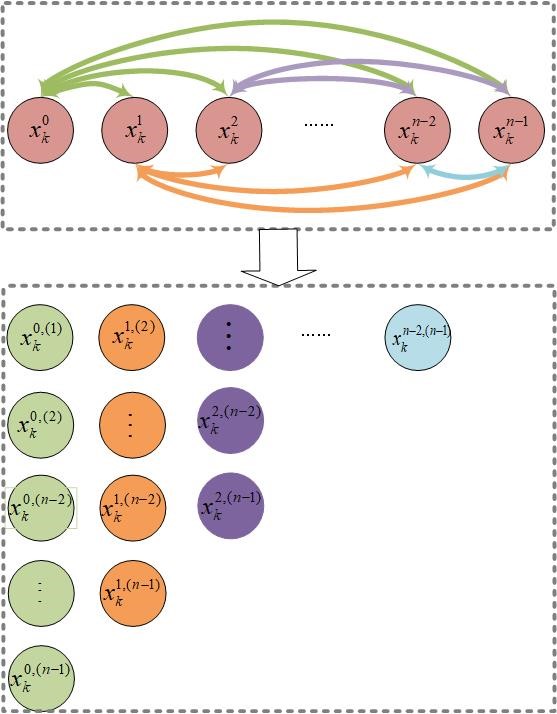}
	\caption{Schematic diagram of the fusion rules between categories of unbalanced data}
	\label{fig3}
\end{figure*}

The fusion rules are shown in Figure \ref{fig3}, where any one selected sample can only be fused with other selected samples other than itself without repeat, essentially a permutation of the samples. The $x_0^{\left( {i,\left( 0 \right)} \right)} \in {x_{sel}}$ in Figure \ref{fig3} represents the fusion of the sample with subscript ${i_{ \ne 0}}$ and the sample with subscript $0$ in class $0$. If all $n$ samples in a class are fused, $\frac{{n\left( {n - 1} \right)}}{2}$ new data will be generated according to the above rules, but the computational complexity of this model is high because the amount of data is very large at this point. To reduce the computational cost, it is necessary to specify the proportion of intra-class fusion $\delta $, where $\delta = 1 $ means that all samples are fused with each other. If  $\delta  \in \left( {0,1} \right)$, then $n * \delta $ samples are randomly selected within the class for fusion.

Step 2) After intra-class fusion of equal number of samples of each category, a new dataset with $\frac{{n\delta \left( {n\delta - 1} \right)}}{2}$ of each category is obtained, at which time the new data has new features of the current kind of samples, and the generated new data is useful for generating a training set feature base and a test set feature base.

Step 3) The fusion procedure is repeated, with the newly generated data from Step 2 fused again for each class, and then the training set\'s feature base is utilized. The propensity for the fusion of multiple data sets to result in an exponential increase in the volume of data necessitates this strategy, which in turn necessitates an increase in computational resources. The second stage effectively reduces the number of feature bases in the training set, preserving a delicate balance between data features and volume. Assuming that the number of feature bases to be generated for each category is $p$, data with dimension is defined by Equ. \ref{eq2}.

\begin{equation}
	\label{eq2}
	m=\frac{n\delta\left ( n\delta-1 \right ) }{2p}
\end{equation}

Equ. \ref{eq2} are then fused into one feature base by sampling without replacement in each category, and a weighted average method will be used in the frequency domain to calculate the wavelet transform. For category $k$, the operation is performed as in Figure \ref{fig3}. Let $x\left[ {m,n} \right]$ be the input two-dimensional signal; $g\left[ n \right]$ be the low-pass filter that filters out the high-frequency part of the signal and outputs the low-frequency part; conversely, $h\left[ n \right]$ be the high-pass filter that filters out the low-frequency signal and outputs the high-frequency signal, and $ \downarrow 2$ be the downsampling filter factor. The $n$-direction is first processed in high-pass, low-pass and downscaling, which are both defined by Equs.\ref{eq3}-\ref{eq4}.

\begin{equation}
	\label{eq3}
	{v_{1,L}}\left[ {m,n} \right] = \sum\nolimits_{k = 0,...,K - 1} {x\left[ {m,2n - k} \right]} g\left[ k \right]
\end{equation}
\begin{equation}
	\label{eq4}
	{v_{1,H}}\left[ {m,n} \right] = \sum\nolimits_{k = 0,...,K - 1} {x\left[ {m,2n - k} \right]} h\left[ k \right]
\end{equation}

Then the high and low pass and downscaling are applied for both $v_{1,L}\left[ m,n\right]$  and $v_{1,H}\left[ m,n\right]$ along the $m = \left\lfloor {\frac{{n\delta \left( {n\delta  - 1} \right)}}{{2p}}} \right\rfloor$ direction, thus we have Equ.\ref{eq5}

\begin{equation}
	\label{eq5}
	\begin{array}{l}
		{x_{1,LL}}\left[ {m,n} \right] = \sum\nolimits_{k = 0,...,K - 1} {{v_{1,L}}\left[ {2m - k,n} \right]} g\left[ k \right]\\
		{x_{1,HL}}\left[ {m,n} \right] = \sum\nolimits_{k = 0,...,K - 1} {{v_{1,L}}\left[ {2m - k,n} \right]} h\left[ k \right]\\
		{x_{1,LH}}\left[ {m,n} \right] = \sum\nolimits_{k = 0,...,K - 1} {{v_{1,H}}\left[ {2m - k,n} \right]} g\left[ k \right]\\
		{x_{1,HH}}\left[ {m,n} \right] = \sum\nolimits_{k = 0,...,K - 1} {{v_{1,H}}\left[ {2m - k,n} \right]} h\left[ k \right]
	\end{array}
\end{equation}

The subsequent phase entails the reconstruction of the signal, which encompasses the inversion of the wavelet transform. This intricate process involves reconstituting a wavelet-transformed signal to reinstate the original signal's integrity. Within this context, the reconstruction of the discrete original signal is executed via the utilization of the subsequent transformative technique as denoted Equ.\ref{eq6},

\begin{equation}
	\label{eq6}
	X\left( z \right) = \sum\nolimits_{n \in N} {x\left( n \right){z^{ - n}}}
\end{equation}
The low-pass filter of \\
$DWT\{x_i\} = \{x_{1,LL}[m,n],x_{1,HL}[m,n]$,\\
$x_{1,LH}[m,n],x_{1,HH}[m,n]\}$ is $Z$-transformed to $G(z)$, and the high-pass filter to $H(z)$, followed by downsampling via Equ. \ref{eq7}.

\begin{equation}
	\label{eq7}
	\left\{ \begin{array}{l}
		{x_{1,L}}\left( z \right) = \frac{1}{2}*\left[ {X\left( {{z^{{1 \mathord{\left/
								{\vphantom {1 2}} \right.
								\kern-\nulldelimiterspace} 2}}}} \right)G\left( {{z^{{1 \mathord{\left/
								{\vphantom {1 2}} \right.
								\kern-\nulldelimiterspace} 2}}}} \right) + X\left( { - {z^{{1 \mathord{\left/
								{\vphantom {1 2}} \right.
								\kern-\nulldelimiterspace} 2}}}} \right)G\left( { - {z^{{1 \mathord{\left/
								{\vphantom {1 2}} \right.
								\kern-\nulldelimiterspace} 2}}}} \right)} \right]\\
		{x_{1,H}}\left( z \right) = \frac{1}{2}*\left[ {X\left( {{z^{{1 \mathord{\left/
								{\vphantom {1 2}} \right.
								\kern-\nulldelimiterspace} 2}}}} \right)H\left( {{z^{{1 \mathord{\left/
								{\vphantom {1 2}} \right.
								\kern-\nulldelimiterspace} 2}}}} \right) + X\left( { - {z^{{1 \mathord{\left/
								{\vphantom {1 2}} \right.
								\kern-\nulldelimiterspace} 2}}}} \right)H\left( { - {z^{{1 \mathord{\left/
								{\vphantom {1 2}} \right.
								\kern-\nulldelimiterspace} 2}}}} \right)} \right]
	\end{array} \right.
\end{equation}

The wavelet transform used in this paper is bior1.3, where two or more signals are fused by bior1.3 wavelet transform and inverse transform to generate a new signal, with the following rules. The new generated signal is obtained by Equ. \ref{eq8}.  For a given set of signals $X = \left\{ {{x_0},{x_1},......,{x_n}} \right\}$
\begin{equation}
	\label{eq8}
	x_{\text{new}}(t) = \text{IWT}\sum_{i=0}^{K-1} \left\{ \frac{1}{2}x_{i,\text{LL}},\frac{1}{2}x_{i,\text{LH}},\frac{1}{2}x_{i,\text{HL}},\frac{1}{2}x_{i,\text{HH}} \right\}
\end{equation}
where $IWT$ denotes wavelet inverse transform compared to $DWT$, and $x_{new}$ is the new signal generated by the wavelet inverse transform after weighted averaging of multiple signals in the frequency domain.

\subsection{Dataset for performance evaluation}
\begin{figure*}[h]
	\centering
	\includegraphics[width=0.9\linewidth]{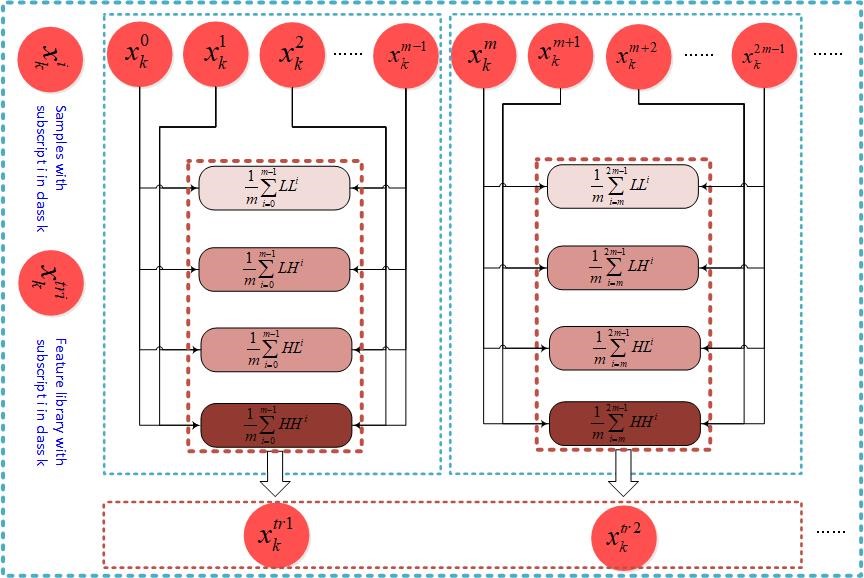}
	\caption{The processing diagram of new training set generation}
	\label{fig4}
\end{figure*}

\begin{figure*}[h]
	\centering
	\includegraphics[width=0.9\linewidth]{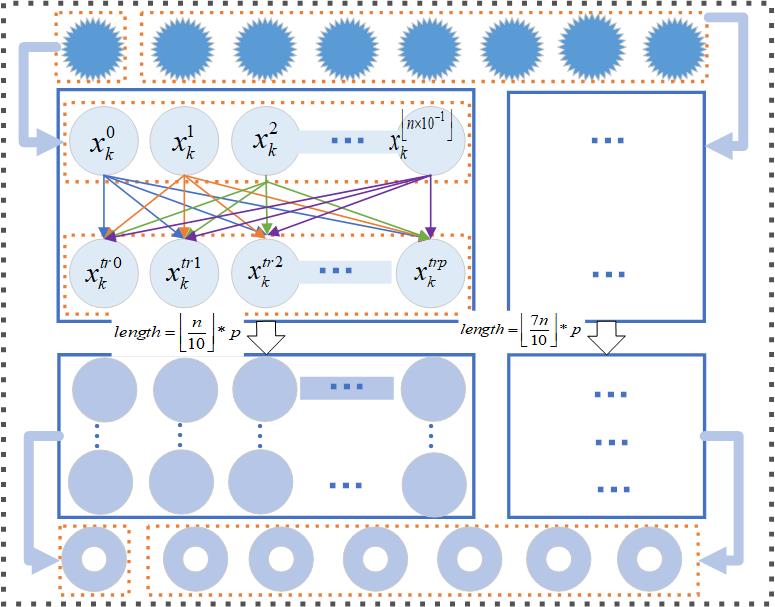}
	\caption{The processing diagram of new validation set generation}
	\label{fig5}
\end{figure*}

The feature bases of the training set, generated in step 3, are fused into a singular feature base, with the fusion process involving a weighted averaging process over the frequency domain, similar to that illustrated in Figure \ref{fig3}. Then, the samples selected by step 2 are divided, as shown in Figure 4, and in the experiment, the training and validation sets account for 80\% and the test set accounts for 20\%. Then we have obtained the training set feature base and test set feature base based on steps 2-3, now we fuse the training set and test set divided with the training set feature base and test set feature base respectively to generate new training data and test data, the detailed process is shown in Figures \ref{fig4}-\ref{fig5}, where $x_{k}^{i}$ ($i$ denotes the sample subscript of class $k$) and $p$ training set feature base fusion will get $p$ training samples.

Note that, the total number of original training sets in each category is $\frac{4}{5}n$, after fusing with the $p$ training set feature libraries that match this class, the resulting training data samples are  $\frac{4}{5}np$. Similarly, each category has only one test set feature library, so the resulting test data sample is $\frac{1}{5}n$.

\subsection{ECG classification methods based on deep learning}
The presence of signal noise in ECGs is inevitable. ECG signals are millivolt-level bioelectric signals that record the depolarization and repolarization of the patient's cardiomyocytes. They are clinically distinguished by their high efficiency, ease, and noninvasiveness. However, during acquisition, ECG signals are subject to various types of noise, such as power frequency interference, myoelectric interference, and baseline drift noise. In order to perform accurate waveform feature extraction and arrhythmia identification of ECG signals, high-quality signals are required to achieve effective arrhythmia detection and categorization.

During acquisition, ECG signals may be susceptible to corruption by noise, such as superimposed white noise, originating from high-frequency interference like myoelectricity and rapid breathing by the patient. In the realm of supervised learning, the precision of label data corresponding to training data plays a pivotal role in achieving optimal learning outcomes. Ensuring accurate annotations or classifications for the training data is essential. Furthermore, deep learning neural networks, renowned for their complexity, necessitate a substantial volume of labeled training data, often referred to as big data or massive data, to yield favorable learning outcomes. It's important to note that a large dataset helps mitigate overfitting, a common challenge in deep learning.

When dealing with learning rates in the context of neural networks, a high learning rate causes the model to initially focus on learning feature information from effective samples and later incorporate information from 'dirty' data. Here, 'effective samples' refer to relevant, informative data points. To stabilize the model's learning process and reduce the influence of 'dirty' data, lowering the learning rate is advisable. However, even with a reduced learning rate, the impact of 'dirty' data on the model's performance cannot be entirely neglected.

In practice, it is recommended to use a slower learning rate when training on datasets containing 'dirty' data, all while ensuring the accuracy of the validation set. If an observation reveals a continuous increase in the recognition rate on the training set coupled with a decline on the validation set, there is a high likelihood that the model has assimilated information from the 'dirty' data. In such cases, it is prudent to halt training to mitigate the risk of $overfitting$ and preserve the model's generalizability.

Although filtering (Wave filtering) can filter out the frequency of a specific band in the signal, the ECG signal typically requires the signal to be as free from distortion as possible, and the filter filters or weakens the unwanted components, invariably accompanied by distortion. The conventional ECG method is known to be sensitive to noise and heavily reliant on filters, which may result in filtered signals that cannot accurately reflect the true ECG signal characteristics. To address the issue of ECG signal distribution imbalance, this paper proposes a data fusion method using wavelet analysis. This approach can effectively suppress the drastic changes in the input caused by noisy samples in the training set and reduce the training time by avoiding the participation of all samples each time the connection weight is adjusted. Deep belief networks can be used as both generative and discriminant models. When used as a generative model, the network generates training data based on a probability distribution. However, achieving a probability distribution covering all data patterns is difficult.

Short cardiac beats, as applied in ECG classification using Convolutional Neural Networks (CNNs), involve the use of truncated segments or fragments of the ECG signal instead of the entire recording. This approach focuses on capturing local features and patterns associated with distinct cardiac events, optimizing the model's ability to discern between various classes. Complementary to this, the proposed data fusion method serves to effectively address challenges related to employing the complete ECG record as input. By strategically combining information from multiple sources, the data fusion method aims to reduce the impact of noise in ECG data. This not only streamlines the model's structure by simplifying the computational requirements but also enhances the accuracy of ECG classification by prioritizing relevant local features within short cardiac beats. The synergy of short cardiac beats and data fusion presents a refined approach, offering improved noise resilience, model simplicity, and classification precision in the analysis of ECG signals.

\section{Experiments}
In this section, we present a thorough description of the experimental data and outline the methodology used for sample selection. Additionally, we provide a comprehensive account of the experiment's implementation and detail the operations performed. Subsequently, we report the experimental results and conclude with a detailed analysis of the findings. The primary data used in this study is sourced from the China Physiological Signal Challenge 2018 (CPSC2018).

"\textbf{A stone from another mountain can be used to polish jade.}" This means that one can learn from the experience and wisdom of others to improve oneself or achieve greater success. To supplement the data for categories with limited samples, we incorporated data from other relevant datasets. The data distribution and sources are presented in Table \ref{tab2}. Notably, the Normal category exhibited the highest number of data samples, which was 5.8 times that of the second-ranked RBBB and 46.6 times that of the category with the lowest number of samples, STE. To mitigate the potential impact of class-imbalance on the model, we employed a standard sample size of 213 for the STE category. Specifically, only 213 samples from each category were used as the original data for the experiment. For categories with a larger number of samples, we randomly selected 213 samples.

\begin{table*}
	\centering
	\caption{Data sources and distribution}
	\label{tab2}
	\begin{tabular}{lcccccc}
		\toprule
		Dataset&CPSC2018&G12EC&PTB-XL&Chaman-Shaoxing&Ningbo&Total\\
		\midrule
		AF&1044&77&35&422&0&1578\\
		I-AVB&722&64&1&0&1&788\\
		LBBB&190&25&13&0&0&228\\
		PAC&638&3&0&0&1&642\\
		PVC&748&0&0&0&0&748\\
		RBBB&1629&72&0&0&0&1701\\
		STD&824&2&0&0&0&826\\
		STE&209&4&0&0&0&213\\
		Normal&981&1752&1287&1366&4542&9928\\
		Total&6985&1999&1336&1788&4544&16652\\
		\bottomrule
	\end{tabular}
\end{table*}

\subsection{Experiment results of the class-imbalance ECG}

Currently, a prevalent method in ECG classification and discrimination involves the use of Convolutional Neural Networks (CNNs). These models typically take individual heartbeats or short segments, such as 3s or 5s, as inputs. However, it is rare to observe studies that utilize the entire ECG recording as input. The utilization of the entire ECG recording would result in a significantly larger input size, leading to an increase in the amount of information and variability of features. To maximize the amount of information extracted, each layer of the model would require more convolutional kernels. However, this would lead to an increase in the overall network size, impeding the focus on local features that can indicate abnormalities in ECG. Therefore, there exists a trade-off between the quantity of information extracted and the ability to focus on localized features.

1) In order to ensure consistency of sample size and comprehensiveness of information, the data collection process for the CPSC2018 dataset entailed recording durations ranging from 6 to 60 seconds. For each segment of the recorded data, we analyzed the mean of the corresponding Gaussian distribution and determined it to be in proximity to 5000. Subsequently, we processed the data utilizing the method detailed in section 2.1, and ultimately reshaped each sample to a format of (12, 5000).

2) In the course of this experiment, we employed a Windows 10 system equipped with a 5960X CPU, 32GB of RAM, and a Nvidia Tesla K80 GPU 24G. The experimental setup for each model employed a batch size of 10, consisting of 150 epochs, 50 steps per epoch, and input samples with a shape of (12, 5000). The learning rate was set at 0.001, while the loss function was specified as cross-entropy, with the Adam optimizer utilized for optimization purposes.

3) In the course of our experiment, we selected VGG16, LeNet5, Inception, and LSTM as our training models. VGG16 is a classic convolutional neural network model that features a deep structure, enabling it to extract higher-level features. This feature is particularly beneficial in the analysis of electrocardiogram (ECG) signals, given the potential presence of various frequencies and amplitudes within these signals. VGG16 can effectively extract these features from raw signals and encode them into higher-level representations, facilitating the identification of different heart diseases or abnormalities. LeNet5, on the other hand, is a relatively straightforward convolutional neural network model that boasts a shallow structure, thereby promoting efficiency in terms of training speed and computational resources. The use of convolutional neural networks can be particularly advantageous when analyzing ECG signals, given their high dimensionality and complexity, as these networks can effectively extract essential features and facilitate further signal processing. Inception, meanwhile, can simultaneously extract features from different time scales, rendering it particularly useful for capturing diverse features present in ECG signals, which may appear at various time scales. Finally, LSTM is a recurrent neural network model that can process sequence data, such as time-series data. In the context of ECG signal analysis, LSTM can take input signals arranged in time order and extract time-dependent features from them. Given the temporal nature of ECG signals, LSTM can capture their temporal characteristics effectively.

4) In this experiment, a comprehensive set of evaluation metrics was employed to assess the performance of the models. Accuracy, loss, recall, precision, receiver operating characteristic (ROC), and area under the curve (AUC) were selected as the evaluation metrics for this study. To validate the robustness of the algorithm, we introduced different types of noise, including BW, EM, and MA, to the samples at various noise intensities. We added 20 noise intensities to the samples, ranging from -27dB to 12dB, including -27dB, -18dB, -12dB, -10dB, -9dB, -8dB, -7dB, -6dB, -5dB, -4dB, -3dB, -2dB, -1dB, 0dB, 1dB, 2dB, 3dB, 6dB, 9dB, and 12dB. We then utilized this data for testing to evaluate the performance of the algorithm under different noise conditions.

\subsection{Model robustness performance verification and result comparison}
\begin{figure*}[h]
	\centering
	\includegraphics[width=\linewidth]{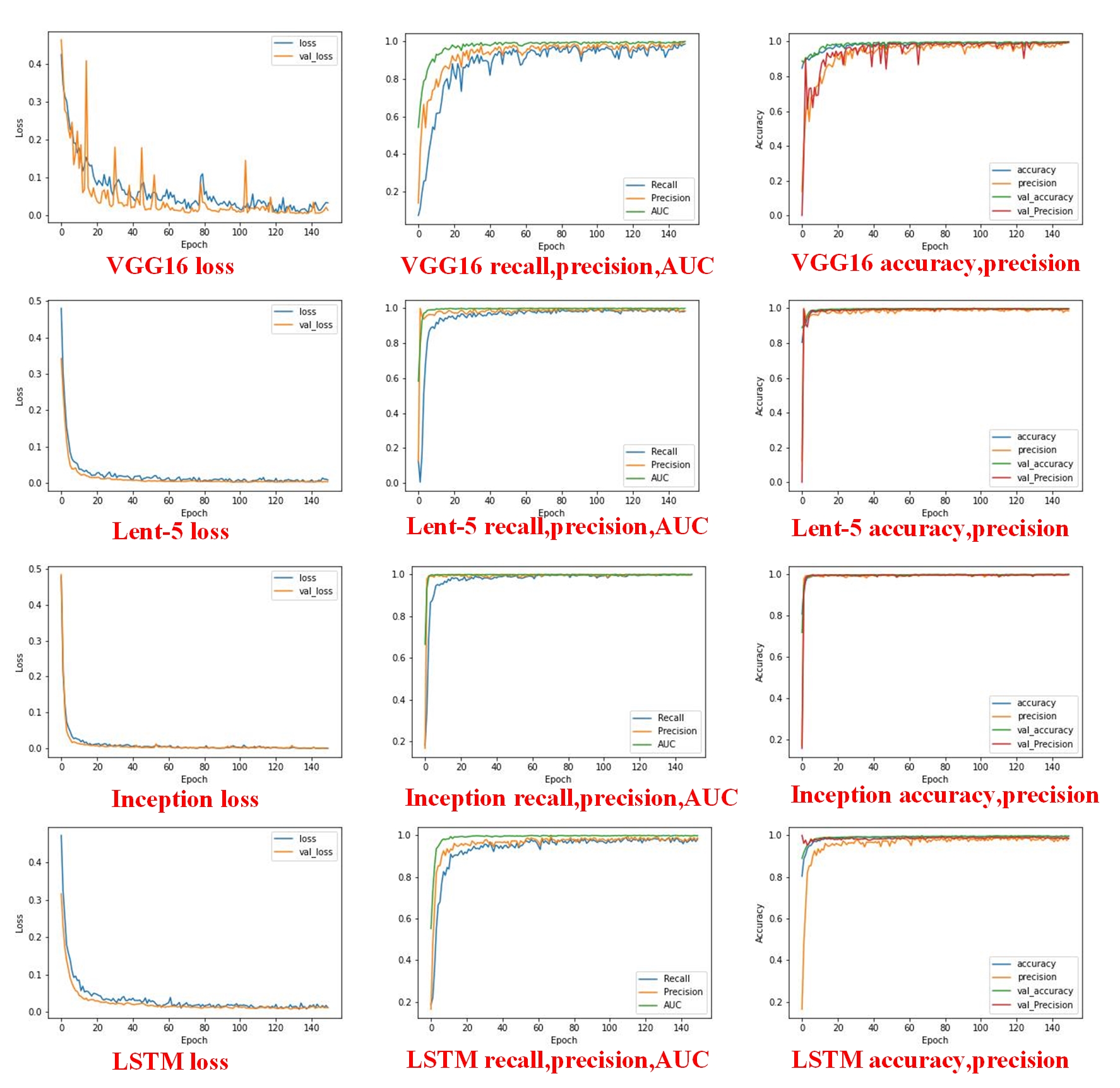}
	\caption{Metrics on the training and validation sets for VGG16, LeNet5, Inception, and LSTM models}
	\label{fig6}
\end{figure*}

\begin{figure*}
	\centering
	\includegraphics[width=\linewidth]{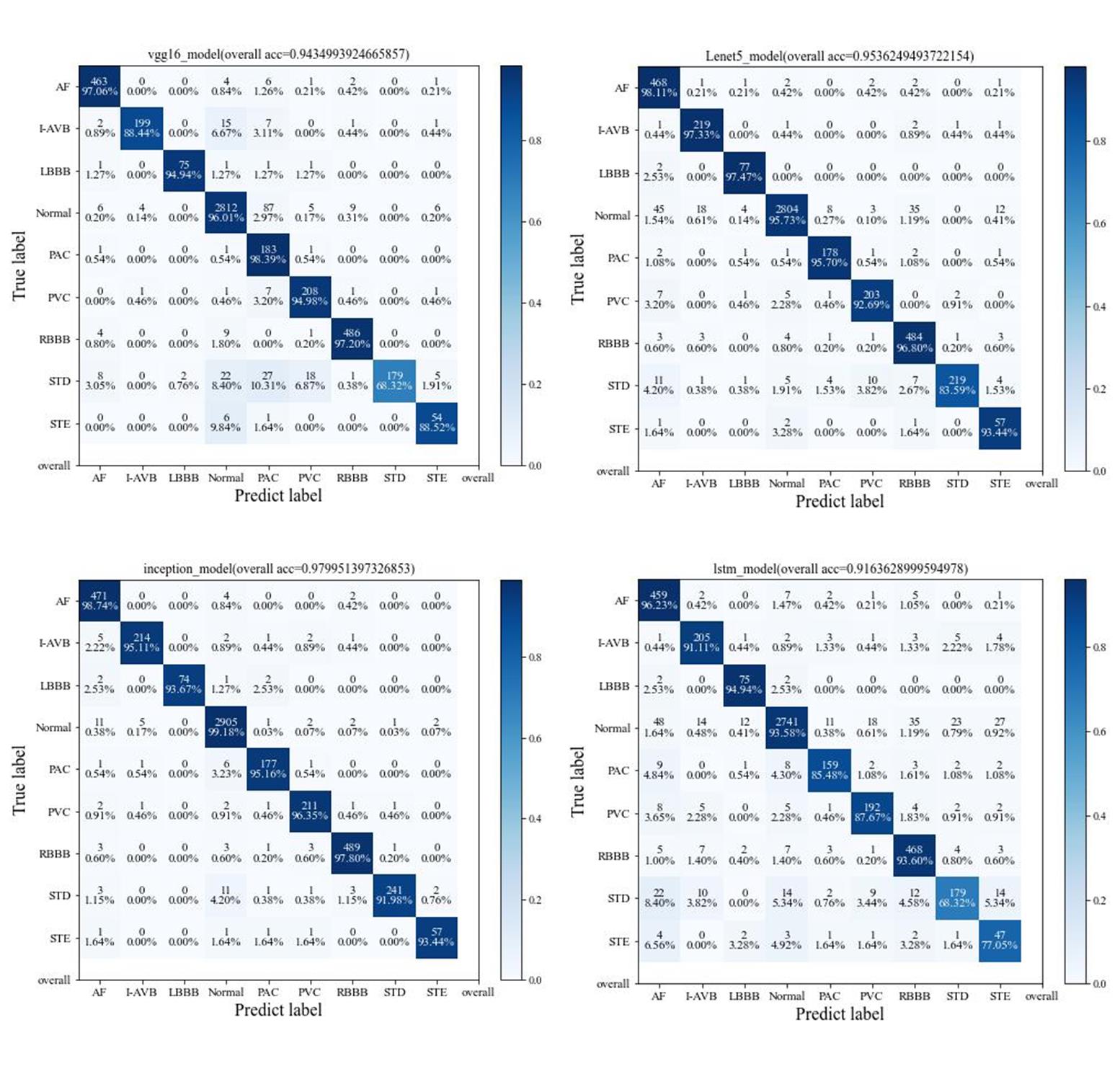}
	\caption{Confusion matrix on the test set for VGG16, LeNet5, Inception, and LSTM models}
	\label{fig7}
\end{figure*}

\begin{figure*}
	\centering
	\includegraphics[width=\linewidth]{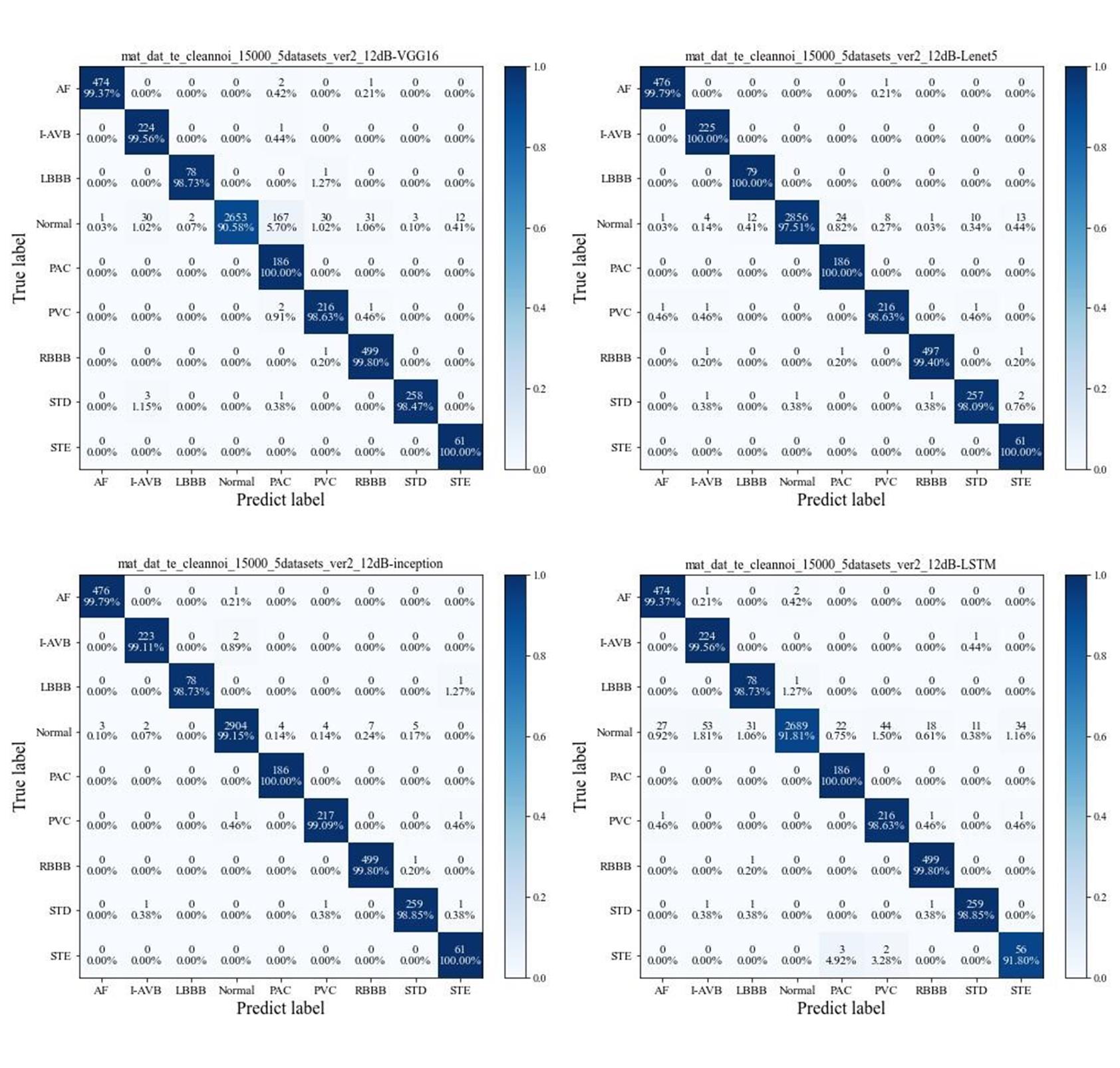}
	\caption{The confusion matrix of the testing of data with 12dB noise added in the VGG16, LeNet5, Inception, and LSTM models}
	\label{fig8}
\end{figure*}

\begin{figure*}
	\centering
	\includegraphics[width=\linewidth]{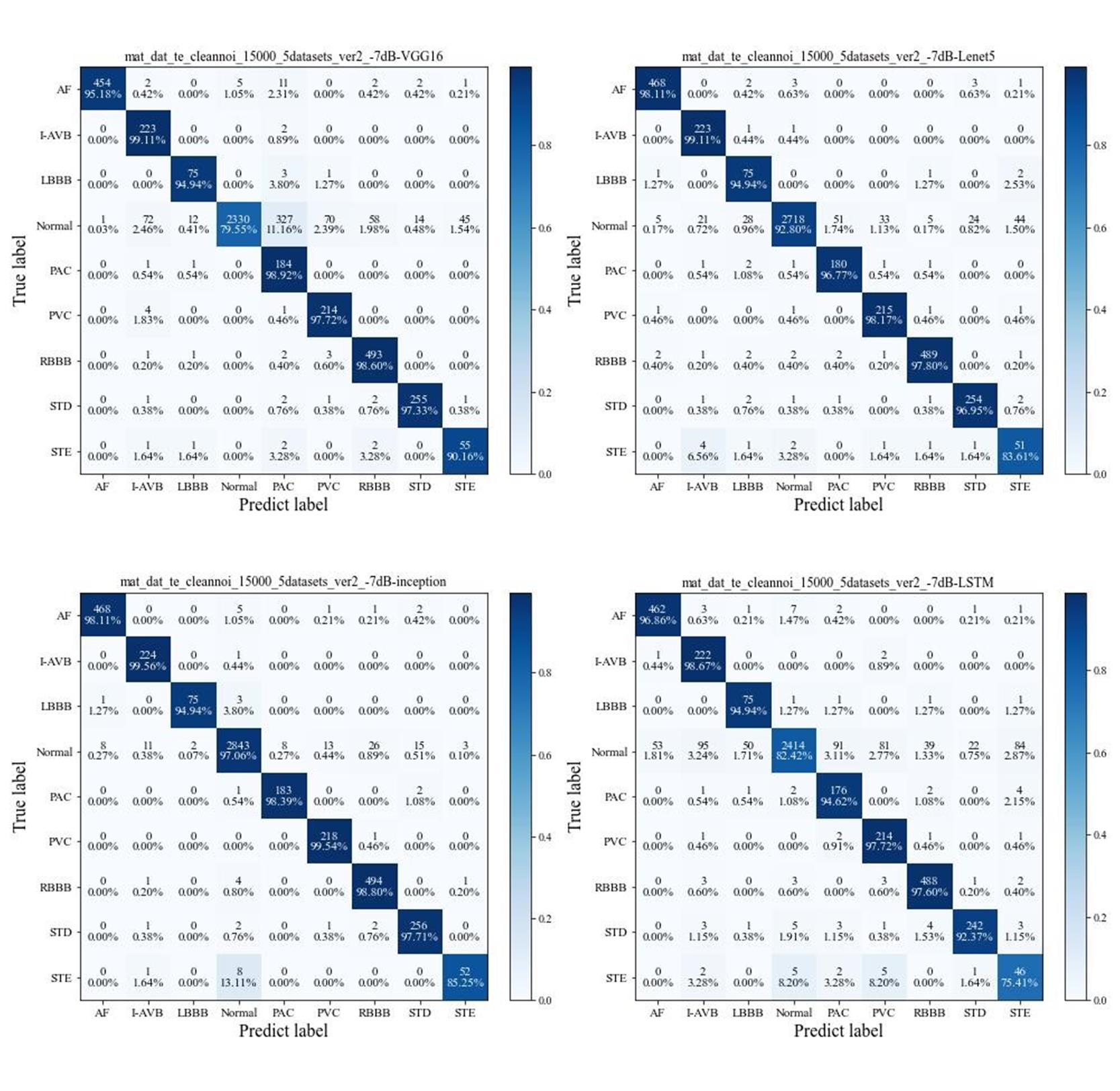}
	\caption{The confusion matrix of the testing of data with -7dB noise added in the VGG16, LeNet5, Inception, and LSTM models}
	\label{fig9}
\end{figure*}

The performance evaluation results of VGG16, LeNet5, Inception, and LSTM models on both training and validation sets are presented in Figure \ref{fig6}. Notably, LeNet5, Inception, and LSTM models exhibit remarkable stability, with accuracy, precision, recall, and AUC metrics close to 1 and losses close to 0 across all models. These findings indicate the effectiveness of the models. The confusion matrices of the test sets for VGG16, LeNet5, Inception, and LSTM models are presented in Figure \ref{fig7}, while Table \ref{tab2} reports the accuracies of the test sets for these models. These results provide valuable insights into the performance of each model on the test sets. Moreover, the confusion matrix for the testing of data with 12dB and -7dB noise added using the VGG16, LeNet5, Inception, and LSTM models under this algorithm are presented in Figures \ref{fig8}-\ref{fig9}, respectively. As the noise increases, there is a decrease in the accuracy of the models.

\section{Experimental Results}
In our experimental efforts to tackle imbalanced ECG data, a nuanced feature fusion approach proved instrumental. Traditionally, addressing imbalanced datasets involves oversampling minority class samples, a practice that can lead to model overfitting and weight degradation, thus compromising performance. Additionally, simply supplementing random samples from underrepresented classes fails to effectively enhance the discriminative power of models for rare cardiac conditions. As evidenced in Table \ref{tab3}, our innovative feature fusion technique emerged as a powerful solution, delivering substantial performance improvements across all evaluated models, with Inception exhibiting the most pronounced enhancement.

The results underscore the remarkable effectiveness of the proposed feature fusion method, particularly in boosting Inception's classification performance. This technique not only mitigated the adverse impacts of imbalanced data but also demonstrated good versatility across different model architectures. Notably, Inception achieved the optimal performance across most categories: it recorded the highest accuracy values for STE, STD, RBBB, PVC, Normal, and AF, and attained a favorable accuracy of 0.95 for I-AVB, LBBB, and PAC.

While our findings validate the success of the proposed method, it is crucial to interpret the overall accuracy with caution, given the class distribution characteristics of the dataset. The Normal class dominates the dataset, accounting for 59.6\% of the total samples, which can inflate the overall accuracy metric. Therefore, evaluating the performance of each individual class is imperative for a comprehensive assessment of model efficacy in clinical scenarios.

Moreover, although LeNet-5, Inception, and LSTM exhibited consistent performance trends across accuracy, precision, loss, recall, and AUC metrics, the imbalanced nature of the data is still reflected in the relatively suboptimal performance of specific rare classes (e.g., STD and STE). This observation highlights the necessity of a holistic evaluation approach that incorporates class-specific metrics, which can provide more meaningful insights into model performance for clinical applications. Our feature fusion approach effectively addresses the challenges posed by imbalanced ECG data and demonstrates significant potential to enhance classification performance, especially for complex models like Inception. The detailed analysis of class-specific metrics also adds depth to the evaluation process, offering valuable guidance for future method improvements and practical applications in ECG signal classification.

In conclusion, our proposed feature fusion approach achieves promising performance across all nine ECG categories evaluated in this study. As the Chinese proverb goes, "A stone from another mountain can be used to polish jade," which means learning from external sources can facilitate self-improvement. Drawing inspiration from this idea, our method leverages features from diverse datasets to supplement the limited samples of underrepresented classes in the target dataset. This effectively addresses the core challenge of imbalanced ECG data, which often impairs the efficiency and robustness of algorithms in automated cardiovascular diagnostic processing and interpretation.

\begin{table}[htbp]
	\centering
	\caption{Data sources and distribution}
	\label{tab3}
	\setlength{\tabcolsep}{6pt} 
	\begin{tabular}{lcccc}
		\toprule
		Model & VGG16 & LeNet-5 & Inception & LSTM \\
		\midrule
		AF(0)     & 0.97 & 0.98 & 0.99 & 0.96 \\
		I-AVB(1)  & 0.88 & 0.97 & 0.95 & 0.91 \\
		LBBB(2)   & 0.95 & 0.97 & 0.95 & 0.95 \\
		Normal(3) & 0.96 & 0.96 & 0.99 & 0.94 \\
		PAC(4)    & 0.98 & 0.96 & 0.95 & 0.85 \\
		PVC(5)    & 0.95 & 0.93 & 0.96 & 0.88 \\
		RBBB(6)   & 0.97 & 0.97 & 0.98 & 0.94 \\
		STD(7)    & 0.68 & 0.84 & 0.92 & 0.68 \\
		STE(8)    & 0.88 & 0.93 & 0.93 & 0.77 \\
		All       & 0.94 & 0.95 & 0.98 & 0.92 \\
		\bottomrule
	\end{tabular}
	\footnotesize \textit{Note}: The values represent the classification accuracy of each model for different ECG categories.
\end{table}

Importantly, we acknowledge the disproportionate representation of infrequently encountered cardiac conditions in existing datasets, underscoring the limitations of conventional techniques like algorithmic generation and oversampling. To overcome these challenges, our paper introduces a groundbreaking ECG classifier that not only tackles class imbalance but also addresses noise-related complexities inherent in ECG data processing.

The key innovation lies in our application of feature fusion based on the wavelet transform, specifically emphasizing wavelet transform-based interclass fusion. By constructing comprehensive training and test feature libraries through this fusion approach, we create more balanced datasets. Notably, when integrated into our ECG model, this methodology yields remarkable recognition accuracies, \textbf{reaching up to 99\% for Normal, 98\% for AF, 97\% for I-AVB, 98\% for LBBB, 96\% for RBBB, 92\% for PAC, 93\% for PVC, 98\% for STD, and 92\% for STE. The average recognition accuracy for these categories consistently ranges between 92\% and 98\%.}

Crucially, our proposed data fusion methodology outshines existing algorithms, setting a new standard for ECG classification accuracy within the CPSC 2018 dataset. This achievement underscores the significance of our approach in advancing the state-of-the-art in ECG analysis, providing a robust solution to class imbalance and noise-related challenges. The groundbreaking recognition accuracies achieved across all categories firmly establish our proposed approach as a pioneering and highly effective method in the field.

\section{Conclusions}
This paper presents an innovative approach to addressing the issue of class imbalance using wavelet transformation. Specifically, the method entails the selection of a category characterized by a scarcity of samples, establishing the number of samples as the threshold, and subsequently eliminating samples from other categories that possess the same number and threshold. The resulting set of screened samples is subjected to two consecutive amalgamations to generate the feature library for the training set. Subsequently, the feature library of the training set undergoes frequency domain averaging to yield the feature library for the test set. The original training and test sets are then merged with their corresponding feature libraries, resulting in novel training and test sets that effectively expand the sample count while minimizing feature loss. Empirical findings evince that this approach yields substantial enhancements in the classification performance of benchmark models such as VGG16, LeNet-5, Inception, and LSTM, attaining maximum recognition accuracies of 99\%, 98\%, 97\%, 98\%, and 96\% for RBBB(6), PAC(4), PVC(5), STD(7), and STE(8), respectively, based on model test results obtained from the publicly available dataset CPSC2018. Moreover, the suggested methodology demonstrates substantial noise resilience characteristics and attains remarkable precision in the classification of electrocardiograms (ECGs) relative to the current classification outcomes obtained from the CPSC 2018 dataset. Notably, it achieves an average recognition accuracy of 95\%, 94\%, 98\%, and 92\% across the eight categories encompassed within the dataset. In essence, this investigation presents a propitious resolution to the predicament of imbalanced data samples in ECG classification, with prospective implications for the progression of more effective diagnostic instruments and therapeutic strategies targeting cardiac pathologies. Furthermore, it is worth emphasizing that this approach can be extrapolated to address physiological datasets exhibiting analogous class imbalance phenomena.

\section{Acknowledgements}
This research is supported by National Natural Science Foundation of China (No. 61902158).

\section{Conflicts of interest}
The authors declare that they have no conflicts of interest to report regarding the present study.

\section{Data availability}
This paper uses public datasets \cite{ref5} for experiments.

\section{Code Deployment and Accessibility Commitment}
All code pertaining to this paper has been deployed on GitHub, \url{https://github.com/Harmenlv/ECG_CPSC_2018}. The open-source code is designed to provide practical reference and guidance for clinical ECG analysis applications, with detailed documentation and reproducible experimental pipelines to facilitate adoption by researchers and clinical practitioners alike. Even though the performance on the current dataset may not fully meet the stringent requirements of real-world clinical scenarios, the core feature fusion methodology presented in this work can be integrated with multi-source clinical experience, domain knowledge, and heterogeneous ECG datasets to further enhance its applicability. We anticipate that this open resource will foster collaborative improvements in imbalanced ECG data processing, and contribute to the development of more robust and clinically relevant automated cardiovascular diagnostic systems.

\bibliographystyle{unsrt}
\bibliography{cas-refs.bib}

\end{document}